\documentclass{article}

\usepackage[dvipsnames]{xcolor} 
\usepackage{xurl}  
\usepackage{graphicx} 
\usepackage[square,numbers]{natbib} 
\usepackage[colorlinks=true,
            linkcolor=red,
            urlcolor=blue,
            citecolor=OliveGreen]{hyperref} 
\usepackage{doi} 
\usepackage{multirow}  
\usepackage{authblk}  
\usepackage{pdflscape}  
\usepackage{afterpage}  
\usepackage[a4paper, total={14.5cm, 24cm}]{geometry}  
\usepackage{tikz}  
\usepackage{booktabs}  

\title{Predicting Fetal Outcomes from Cardiotocography Signals Using a Supervised Variational Autoencoder}
\author{John Tolladay}
\author{Beth Albert}
\author{Gabriel Davis Jones}
\affil{Oxford Digital Health Labs, Nuffield Department of Women's and Reproductive Health, University of Oxford, UK}

\date{September 2025}

\begin{document}
\bibliographystyle{unsrt}  
\maketitle

\begin{abstract}
\noindent\textbf{Objective:}
To develop and interpret a supervised variational autoencoder (VAE) model for classifying cardiotocography (CTG) signals based on pregnancy outcomes, aiming to address the interpretability limitations of current deep learning approaches.\\
\textbf{Methods:}
The OxMat CTG dataset was used to train a VAE on five-minute fetal heart rate (FHR) segments, labeled with postnatal outcomes. The model was optimised for signal reconstruction and outcome prediction, incorporating Kullback–Leibler divergence and total correlation (TC) constraints to structure the latent space. Performance was evaluated using area under the receiver operating characteristic curve (AUROC) and mean squared error (MSE). Interpretability was assessed using coefficient of determination, latent traversals and unsupervised component analyses.\\
\textbf{Results:}
The model achieved an AUROC of 0.752 at the segment level and 0.779 at the CTG level, where predicted scores were aggregated. Relaxing TC constraints improved both reconstruction and classification performance. Latent analysis showed that baseline-related features (e.g., FHR baseline, baseline shift) were well represented and aligned with model scores, while other metrics like short- and long-term variability were less strongly encoded. Traversals revealed clear signal changes for baseline features, while other signal properties were entangled or subtle. Unsupervised latent decompositions corroborated these patterns.\\
\textbf{Findings:}
This work demonstrates that supervised VAEs can achieve competitive fetal outcome prediction while partially encoding clinically meaningful CTG features. The irregular, multi-timescale nature of FHR signals poses challenges for disentangling physiological components, distinguishing CTG from more periodic signals such as ECG. Although full interpretability was not achieved, the model supports clinically useful outcome prediction and provides a foundation for future interpretable, generative models.

\end{abstract}


\section{Introduction}

Cardiotocography (CTG) is the primary tool for fetal monitoring, recording fetal heart rate (FHR) and uterine activity (UA) to support antepartum assessment and guide intrapartum care. Clinicians use CTGs to detect fetal distress and trigger interventions aimed at preventing adverse pregnancy outcomes such as neonatal acidosis or hypoxia \cite{alfirevic17continuous, mbarek23computerized}. However, CTG interpretation has remained highly subjective, with considerable inter- and intra-observer variability \cite{trimbos78observer, rei16interobserver}. Even experienced obstetricians frequently disagree on CTG trace classification, and the low specificity of CTG interpretation can lead to unnecessary interventions (including caesarean deliveries) without a corresponding reduction in poor neonatal outcomes \cite{alfirevic17continuous}. These limitations have long motivated efforts to develop automated, objective approaches to CTG interpretation.

Computerised techniques have been developed to support CTG analysis, aiming to improve detection and prediction of fetal compromise. Traditional approaches have used hand-engineered features from the FHR and UA signals such as baseline, variability, accelerations and decelerations. These reflect clinical parameters defined in guidelines like FIGO \cite{ayres2015figo} and NICE \cite{nice2024website}. Classical machine learning models classify fetal status from these pre-processed features, often using curated datasets with expert-defined labels (e.g., umbilical artery pH or Apgar score) \cite{jones2025drperformance, jones2022computerized, mendis23review-computerised}. While effective in controlled settings, they depend on manual feature selection and show inconsistent performance across datasets. More recently, deep learning methods have been applied to raw CTG signals, with convolutional and recurrent neural networks achieving state-of-the-art results \cite{xie24review-ai}. Their clinical adoption is hindered by poor interpretability, limited data and variable generalisability. The ``black box'' nature of these models raises concerns about trust and explainability, highlighting the need for methods that combine predictive power with interpretability. Deep generative models, such as supervised variational autoencoders (VAEs), offer a potential solution by enabling outcome prediction alongside structured representation of learned features.

VAEs have been successful at distinguishing CTG segments labeled as ``suspicious'', ``pathological'' or ``normal'' by majority voting of a panel of three expert clinicians, achieving an area under the receiver operating characteristic curve of 0.94--0.96 when distinguishing ``normal'' from ``pathological'' segments \cite{melaet24seqvae, vries25seqvae}. Similar models have also been effectively applied to model single- and multi-beat ECG signals, enabling the generation of realistic signals via latent variables that capture meaningful physiological features \cite{jang21unsupervised, beetz22multidomain, patika24artificial, harvey2024comparison}. Motivated by the potential of these models, this study explores a supervised VAE for predicting fetal outcomes from FHR segments. The approach combines the feature-learning capability of deep neural networks with the structured, interpretable latent space of probabilistic generative models. In the absence of expert-labeled segments, FHR data from the extensive OxMat CTG dataset are labeled according to pregnancy outcomes \cite{khan2024oxmatdatasetmultimodalresource}. Latent space traversals, partial least squares regression and analysis of coefficients of determination are then applied to perform a novel investigation in to the interpretability of the model.


\section{Method} \label{sec:methods}


\subsection{Data Selection and Splitting}

\begin{figure}[htbp]
    \footnotesize
    \textbf{Normal Pregnancy Outcome Group} - To be included in this set, each case was required to meet all of the following conditions:
    \begin{itemize}
        \item Mother:
        \begin{itemize}
            \item Age between 18 and 40 years
            \item Body Mass Index (BMI) $\leq 30$
        \end{itemize}
        \item Baby:
        \begin{itemize}
            \item Gestational age at birth between 37 and 41 weeks
            \item Birthweight between the 10th and 90th percentile
            \item Apgar score $\geq 4$ at 1 minute and $\geq 7$ at 5 minutes
            \item No resuscitation required
            \item No admission to neonatal intensive care
            \item No perinatal infections or respiratory conditions
        \end{itemize}
        \item CTG signal:
        \begin{itemize}
            \item If multiple CTGs were available in the same gestational week, only the \textit{first} trace was used to avoid bias introduced by follow-up monitoring
        \end{itemize}
    \end{itemize}
    
    \noindent\textbf{Adverse Pregnancy Outcome Group} - CTGs were included in this group only if the baby met one or more of the following clinical conditions:
    \begin{itemize}
        \item Intrauterine growth restriction (birthweight $\leq$ 3rd percentile) with Apgar score $<4$ at 1 minute and $<7$ at 5 minutes
        \item Evidence of acidemia:
        \begin{itemize}
            \item \textit{No labour}: arterial pH $<7.13$ and base excess $>10$
            \item \textit{With labour}: arterial pH $<7.05$ and base excess $>14$
        \end{itemize}
        \item Any of the following outcomes:
        \begin{itemize}
            \item Apgar $<4$ at 1 minute and $<7$ at 5 minutes
            \item Stillbirth or death within 24 hours
            \item Neonatal death
            \item Asphyxia
            \item Hypoxic-ischaemic encephalopathy
            \item Neonatal sepsis
            \item Perinatal infection
            \item Respiratory condition
            \item Neonatal intensive care admission lasting 7 days or more
        \end{itemize}
        \item CTG must have been recorded within 7 days of birth, to ensure relevant signal content (which, due to the limit of CTGs only being included where recorded at $<37$ weeks gestation, means all APO cases were born preterm).
    \end{itemize}
    \caption{Details for the conditions used to extract and split cardiotocography signals for the normal and adverse pregnancy outcome groups.}
    \label{fig:data-split-conditions}
\end{figure}

\begin{figure}[htbp]
    \footnotesize
    \centering
    \begin{tikzpicture}[
    node distance=2cm and 4cm,
    every node/.style={draw, rectangle, minimum width=2cm, align=center}
    ]
        \node (head) at (3.5, 0) {208,115 OxMat CTG Records};
        \node (a1) at (0, -1) {15,017 APO CTG Records};
        \node (a2) at (0, -2) {14,807 CTG files of sufficient duration};
        \node (a3) at (0, -3) {228,262 5-minute segments};
        \node (c1) at (7, -1) {10,840 NPO CTG Records};
        \node (c2) at (7, -2) {10,541 CTG files of sufficient duration};
        \node (c3) at (7, -3) {104,078 5-minute segments};
        \node (comb) at (3.5, -4) {332,340 5-minute segments};
        \draw [->] (head)--(a1);
        \draw [->] (a1)--(a2);
        \draw [->] (a2)--(a3);
        \draw [->] (a3)--(comb);
        \draw [->] (head)--(c1);
        \draw [->] (c1)--(c2);
        \draw [->] (c2)--(c3);
        \draw [->] (c3)--(comb);
    \end{tikzpicture}
    \caption{Details of the number of cardiotocography (CTG) signals used and excluded for the normal pregnancy outcome (NPO) and adverse pregnancy outcome (APO) groups}
    \label{fig:data-diagram}
\end{figure}
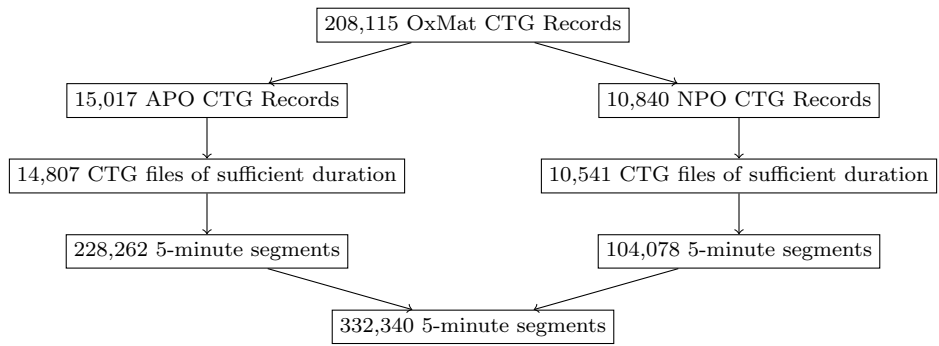

The OxMat dataset comprises 208,115 antepartum cardiotocography (CTG) records collected from 1990 until 2024. It includes extensive demographic and medical data for both mothers and babies at the antepartum, intrapartum and postpartum stages. For this study, two subsets were derived: a normal pregnancy outcome (NPO) group and an adverse pregnancy outcome (APO) group, based on stringent inclusion criteria, as detailed in Figure \ref{fig:data-split-conditions} and validated in a previous study \cite{jones2025drperformance}. All eligible CTG segments meeting inclusion criteria were used and no formal sample size calculation was performed, as the study leveraged the complete available dataset. Only CTGs recorded between 27 and 36 weeks gestation are included within the two datasets (to avoid late-pregnancy influences) and babies in the APO group were all born preterm. CTGs were split in to five-minute segments, overlapping by two and a half minutes with one another. The number of CTGs and FHR segments extracted are detailed in Figure \ref{fig:data-diagram}.


\subsection{Data Pre-processing}

FHR segments with more than 25\% missing data were excluded. Remaining missing values were assigned a placeholder value, while segments between 3.75 and 5 minutes in duration were padded at the end using a separate padding value. Legacy recordings using the older ``epoch'' format (3.75\,s per sample) were converted to $4$\,Hz via up-sampling and signal smoothing. A basic noise reduction algorithm was applied to FHR signals to suppress unnatural outlier values, spikes, maternal heart rate contamination and other common CTG artifacts.

\begin{table}[htbp]
    \centering
    \begin{tabular}{lccccc}
        \toprule
        \textbf{Metric} & \textbf{NPO} & \textbf{APO} & \textbf{Train} & \textbf{Validation} & \textbf{Test} \\
        \midrule
        APO Cases & 0.00 & 100.00 & 68.45 & 68.96 & 68.72 \\
        CTGs & 10,485 & 14,746 & 16,823 & 4,204 & 4,203 \\
        Segments & 104,047 & 227,133 & 220,031 & 55,382 & 55,767 \\
        Male / Female & 0.97 & 1.09 & 1.03 & 1.06 & 1.07 \\
        Gestational Age & 32.67 & 32.35 & 32.48 & 32.53 & 32.47 \\
        & (2.84) & (2.78) & (2.80) & (2.83) & (2.81) \\
        Birthweight & 3280 & 1831 & 2432 & 2437 & 2434 \\
        & (359) & (774) & (954) & (955) & (964) \\
        Missing FHR & 4.16 & 3.75 & 3.87 & 3.87 & 3.95 \\
        \bottomrule
    \end{tabular}
    \caption{Summary statistics for the normal pregnancy outcome (NPO), adverse pregnancy outcome (APO), training, validation and test datasets. The percentage of APO cases within the set, number of cardiotocographs (CTGs) and segments, sex ratio, mean gestational age in weeks and birthweight in grammes (with standard deviations in brackets), and percentage of missing fetal heart rate (FHR) data points are included.}
    \label{tab:dataset_summary}
\end{table}

Before model training the segments are split in to training, validation and test sets. The NPO and APO segments are split separately by their unique CTG identifier, such that segments from a specific CTG all appear within the same set, while maintaining a similar ratio of APO to NPO cases within each set. Table \ref{tab:dataset_summary} summarises the key characteristics of these datasets, including demographic and clinical variables as well as class distribution, offering a comprehensive overview of the data composition. Of the total segments, one sixth are placed in the validation set and one sixth in the test set, with the remaining segments being used for training. The validation set is used to stop the model training before over-fitting occurs, while the test set is used after training to examine the effectiveness of the model on unseen data.

During training, each mini-batch was constructed to contain an equal number of segments from NPO and APO cases in the training set, in order to mitigate the effects of class imbalance. This approach is equivalent to oversampling the minority class so that its frequency matches that of the majority class. Segments with a standard deviation below 1\,bpm or a range (maximum minus minimum) of less than $5$\,bpm were also excluded (1160 segments, $\ll0.1\%$) to prevent any flat-line signals from negatively impacting model performance.


\subsection{Model Architecture and Configuration}

We developed a variational autoencoder (VAE) that jointly reconstructs fetal heart rate (FHR) signals and predicts fetal outcome as a continuous score between 0 (normal) and 1 (adverse). The loss function combines mean squared error (MSE) for signal reconstruction, binary focal cross-entropy for outcome prediction, and the standard $\beta$-TC-VAE terms \cite{chen19tcvae}: a Kullback-Leibler (KL) divergence term weighted by $\beta$ and a total correlation (TC) term weighted by $\lambda$, to promote disentangled and structured latent representations. The $\beta$ and $\lambda$ coefficients are dynamically adjusted during training to constrain KL and TC values within target thresholds. KL Divergence was normalised by dividing the total value by the latent dimension. A target value of $0.5$ (per latent dimension) was then selected to encourage informative latent representations without overpowering the reconstruction loss or collapsing the posterior. Various values were also explored for the total correlation (TC) target, with selection informed by downstream performance and stability considerations.

Model inputs comprise raw FHR signals, with specific values assigned to missing and padding points, along with the corresponding Fast Fourier Transforms (FFTs). A pre-processing layer standardises the FHR signals using the global training set mean and standard deviation, replaces specified missing/padding values with learned tokens and embeds values using a fully connected layer with learned positional encoding. FFT inputs are normalised to the range [0, 1], tokenised similarly and also enriched with positional encoding. These provide further information to the model regarding frequency patterns that might not be obvious from the raw time-domain signal. FHR and FFT embeddings are then passed through separate single-layer, single-head transformers. Their outputs are concatenated and passed through a fully connected layer to define the latent space, parameterised by a mean and log-variance per dimension.

Latent variables are sampled using the reparameterisation trick, drawing from a normal distribution parameterised by the mean and log-variance. These samples are expanded through a fully connected layer and passed through a single-head transformer layer to reconstruct the original FHR signal via a linearly activated output layer. In parallel, a classification head processes the normalised latent variables to provide outcome prediction scores using a sigmoid-activated dense layer, promoting a latent space that encodes discriminative features useful for predicting fetal outcomes. Model hyperparameters were configured to their optimal settings to maximise area under the receiver operating characteristic curve for the predicted scores and minimise reconstruction MSE. This was carried out using a manual search due to the hyperparameter space being impractical to explore completely.


\subsection{Model Performance Metrics} \label{sec:model_performance}

Model performance was evaluated using two metrics: the area under the receiver operating characteristic curve (AUROC) for classification performance and mean squared error (MSE) for reconstruction quality. Model calibration was evaluated using the expected calibration error (ECE), which measures how closely the predicted scores align with the likelihood of the corresponding outcomes. Classification performance was assessed at the segment level, where each FHR segment was evaluated individually against its corresponding label, and at the case level, where predictions for all segments from a single CTG recording were aggregated using the median predicted score to generate a single case-level prediction. The median was selected over other metrics like mean, minimum or maximum as it provided the optimal AUROC. Overfitting was mitigated by separate training/validation/test splits and implementation of early-stopping based on validation loss. To investigate which conditions the model is most applicable to, its ability to predict specific adverse pregnancy outcomes (as detailed in Figure \ref{fig:data-split-conditions}) based on the latent space representations was also compared using AUROC at the segment and case levels.


\subsection{Interpretation Analysis}

To assess how clinically relevant features are represented in the learned latent space, the coefficient of determination ($R^2$) was computed between various features and the latent variables of the trained model. The features included baseline fetal heart rate (FHR), baseline shift (change in baseline between the start and end of each segment), baseline anomaly (deviation from global mean baseline $\approx 140$), short-term variability (STV), long-term variability (LTV), the standard deviation (SD) of the FHR segment, FHR range (maximum - minimum value), and counts of accelerations and decelerations. These features capture various aspects of FHR dynamics that are used clinically to assess fetal well-being. The coefficient of determination was also calculated to assess the correlation between these features and the labels, predicted scores and the errors on these predictions.

We employed partial least squares (PLS) regression to identify latent directions most predictive of each feature. This enabled exploration of the relationships between the features and latent space that goes beyond unsupervised latent analysis. To probe the structure of the latent space, latent traversals were performed: starting from the mean latent vector, shifts along each feature-associated direction were applied over a range of $-10$ to $+10$ standard deviations. Each shifted latent vector was decoded to reconstruct an FHR segment, allowing us to visualise how variation along specific latent directions/dimensions affects the signal. This approach enabled interpretation of how each feature is encoded, providing quantification of its alignment with the latent space (via $R^2$) and visualising its effect on the decoded output. Similarly, traversals along individual latent dimensions were performed to assess the interpretability in isolation. These were applied over a range of $-5$ to $+5$ standard deviations to avoid excessively noisy signals at the extremes. Independent and principal component analyses were also applied to the latent representations to identify the strongest components and visualise how they modify the decoded mean signal. This unsupervised approach allowed us to examine whether traversals along these components corresponded to clinically meaningful or visually appreciable features in the decoded FHR signals.


\section{Results}


\subsection{Model Performance}

\begin{table}[htbp]
    \centering
    \begin{tabular}{ccccc}
        \toprule
        Target TC & Final TC & MSE & AUROC & ECE \\
        \midrule
        3 & 3.127 (0.224) & 20.626 (0.469) & 0.574 (0.005) & 0.272 (0.074) \\
        20 & 19.680 (1.773) & 10.059 (0.368) & 0.681 (0.010) & 0.183 (0.044) \\
        50 & 47.466 (1.666) & 6.669 (0.193) & 0.730 (0.005) & 0.185 (0.060) \\
        200 & 112.210 (1.596) & 5.645 (0.248) & 0.740 (0.009) & 0.183 (0.027) \\
        \bottomrule
    \end{tabular}
    \caption{Mean squared error (MSE) on reconstruction of 5-minute fetal heart rate segments, area under the receiver operating characteristic curve (AUROC), expected calibration error (ECE) and final total correlation (TC) values at the end of training, for a range of target values for the TC of the latent space. Metrics are mean values for the test datasets across 3 models trained using different initialisation seeds, with standard deviations displayed in brackets.}
    \label{tab:total-correlation}
\end{table}

The best performing model configuration was able to distinguish fetal heart rate (FHR) segments between the normal pregnancy outcome (NPO) group and adverse pregnancy outcome (APO) group with an average area under the receiver operating characteristic curve (AUROC) of $0.740$ (standard deviation $\sigma$ = 0.009) and average mean-squared-error (MSE) of $5.645$ ($\sigma$ = 0.248), across three different initialisation seeds. This was achieved using a target total correlation (TC) of $200$, which effectively led to $\lambda=0$ and no constraint on the TC of the latent space. Table \ref{tab:total-correlation} shows example MSE and AUROC values and the TC at end of training for a range of target values. The automated adjustment of $\lambda$ maintains the final TC value very close to the target in all cases except the target value of $200$. In this case the TC grew over the course of training and leveled out at approximately $100$.

\begin{figure}[htbp]
    \centering
    \includegraphics[width=\linewidth]{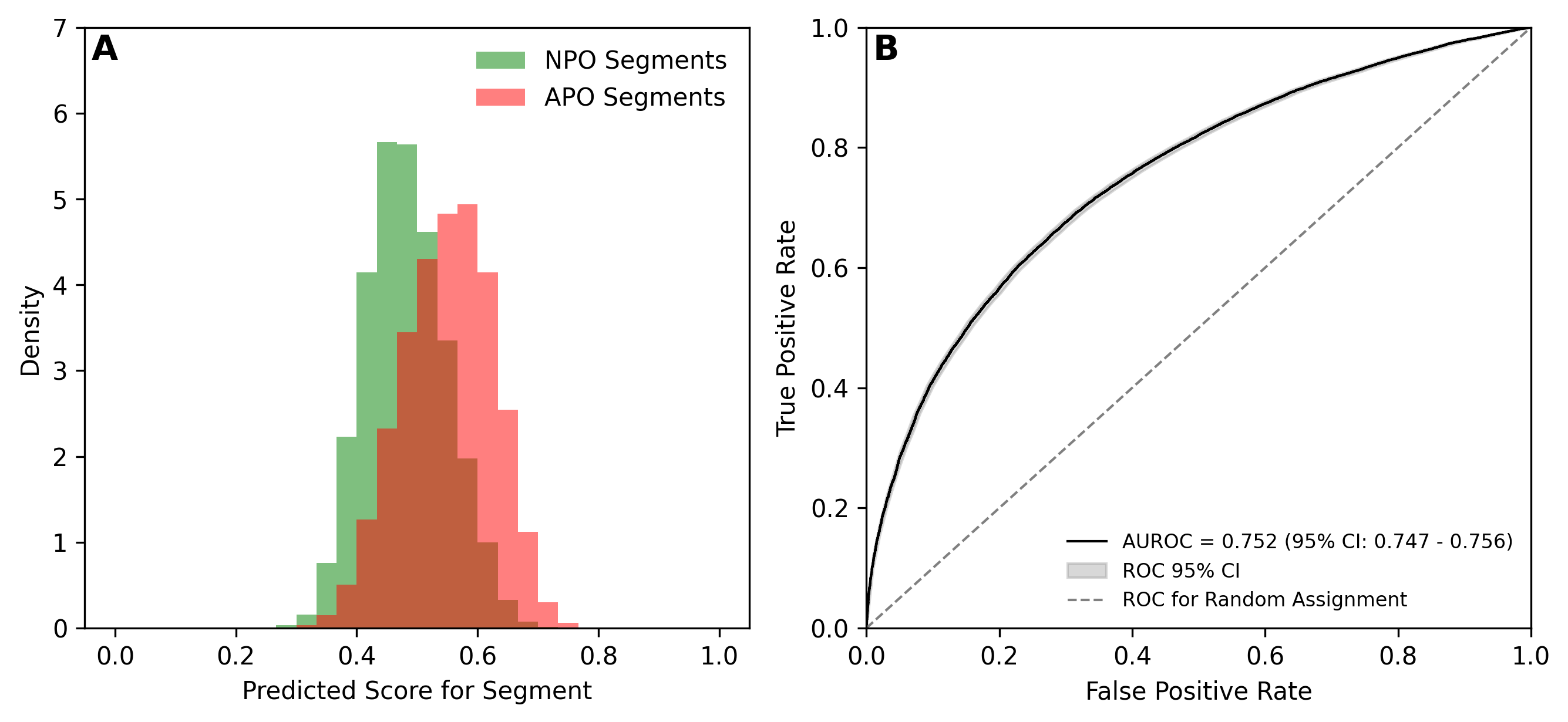}
    \caption{Classification performance of the best performing model when assessing the score assigned to all segments within the test dataset. \textbf{A} shows a histogram of the predicted scores for each segment from the normal pregnancy outcome (NPO) and adverse pregnancy outcome (APO) groups. \textbf{B} shows the receiver operating characteristic curve for the same data with area under this curve (AUROC) of $0.752$ (95\% CI: 0.747--0.756) shown in the legend.}
    \label{fig:segment_predictions}
\end{figure}

\begin{figure}[htbp]
    \centering
    \includegraphics[width=\linewidth]{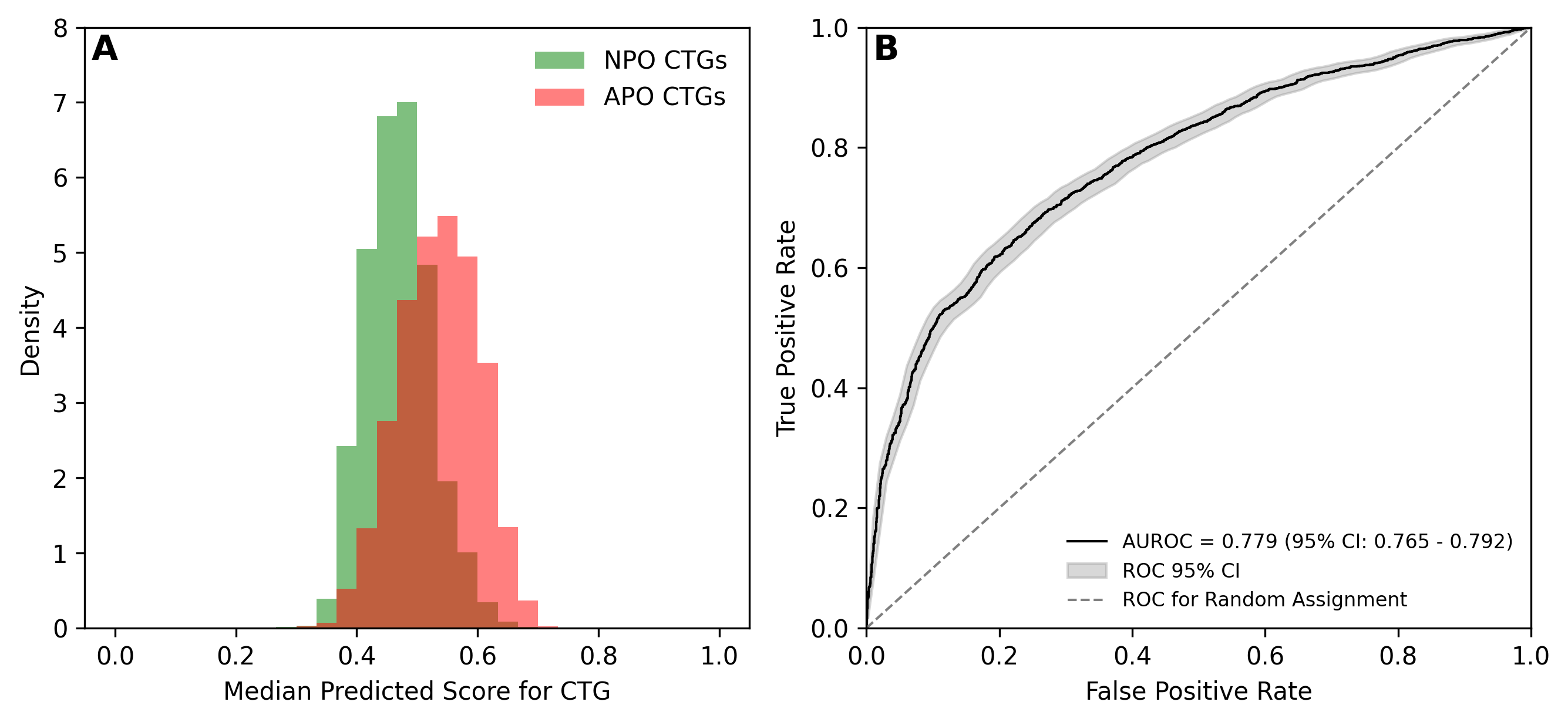}
    \caption{Classification performance of the best performing model when assessing the median score for all overlapping fetal heart rate segments of each cardiotocogram (CTG) in the test dataset, after excluding any segments exceeding the $25\%$ missing data threshold. \textbf{A} shows a histogram of the median predicted scores for both the normal pregnancy outcome (NPO) and adverse pregnancy outcome (APO) groups. \textbf{B} shows the receiver operating characteristic curve for the same data with the area under this curve (AUROC) of $0.779$ (95\% CI: 0.765--0.792) shown in the legend.}
    \label{fig:ctg_predictions}
\end{figure}


The distribution of scores predicted by the best-performing model for NPO versus APO groups is summarised in Figure~\ref{fig:segment_predictions}. While there is substantial overlap between the two distributions, the receiver operating characteristic (ROC) curve shows that the model was able to distinguish the groups with an AUROC of 0.752 (95\% CI: 0.748--0.756) at the individual FHR segment level. At Youden’s threshold (maximising the difference between true positive rate and false positive rate) the model reached a sensitivity of 83.7\% (95\% CI: 83.3–84.1\%) and a specificity of 71.9\% (95\% CI: 71.2–72.5\%), corresponding to an F1 score of 73.7\% (95\% CI: 73.3–74.0\%). When predictions were aggregated by computing the median score across all overlapping segments from a given CTG, the discrimination improved, yielding an AUROC of 0.779 (95\% CI: 0.765--0.793), as shown in Figure~\ref{fig:ctg_predictions}. The sensitivity for aggregated scores remained similar, at 82.6\% (95\% CI: 80.9--84.5\%) while the specificity increased to 82.6\% (95\% CI: 80.9--84.3\%).

\begin{figure}[htbp]
    \centering
    \includegraphics[width=\linewidth]{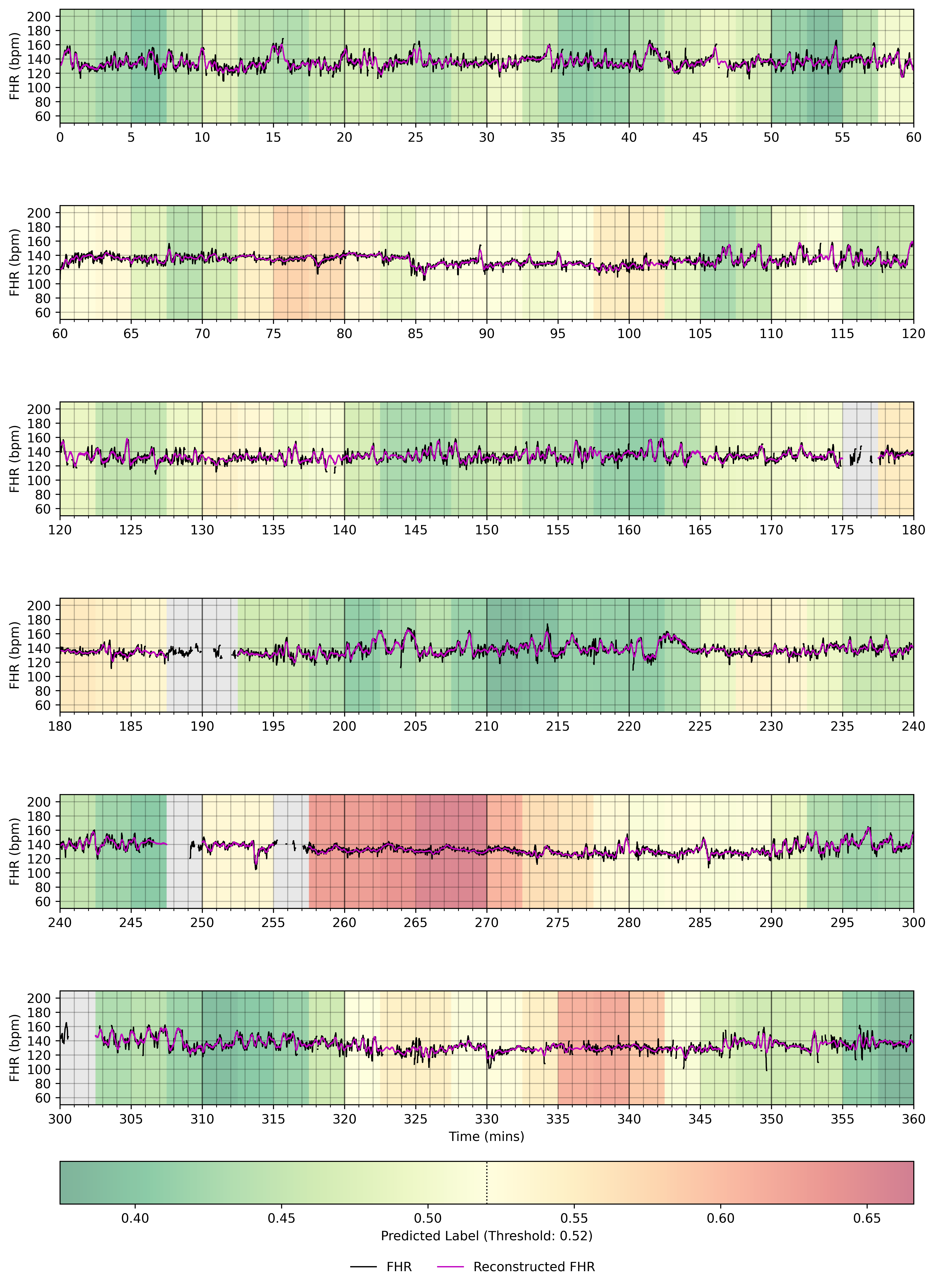}
    \caption{Example model output for a six-hour of fetal heart rate (FHR) recording from the normal pregnancy outcome group. The original FHR signal (black) is overlaid with the model-reconstructed signal (magenta). Background colours indicate the mean score for each 2.5-minute interval, calculated from overlapping five-minute input segments. Grey backgrounds denote intervals with insufficient FHR data for prediction. The colour-bar is centred on Youden's threshold (0.52), determined by maximising the difference between the true positive rate and the false positive rate.}
    \label{fig:output-example}
\end{figure}

A representative six-hour FHR trace from the normal pregnancy outcome group is shown in Figure~\ref{fig:output-example} along with the outputs from the best-performing model. The reconstructed FHR closely follows the raw signal but appears smoother, clearly removing noise spikes in some sections while potentially omitting more subtle variability in others. Predicted scores remain predominantly below the 0.52 decision threshold (green shading), consistent with the normal outcome label, and largely correspond to intervals containing reassuring features such as high variability and accelerations. Other sections, typically with lower baselines, score closer to the threshold (yellow shading), while periods with particularly low variability receive considerably higher scores (red shading).

\begin{figure}[htbp]
    \centering
    \includegraphics[width=\linewidth]{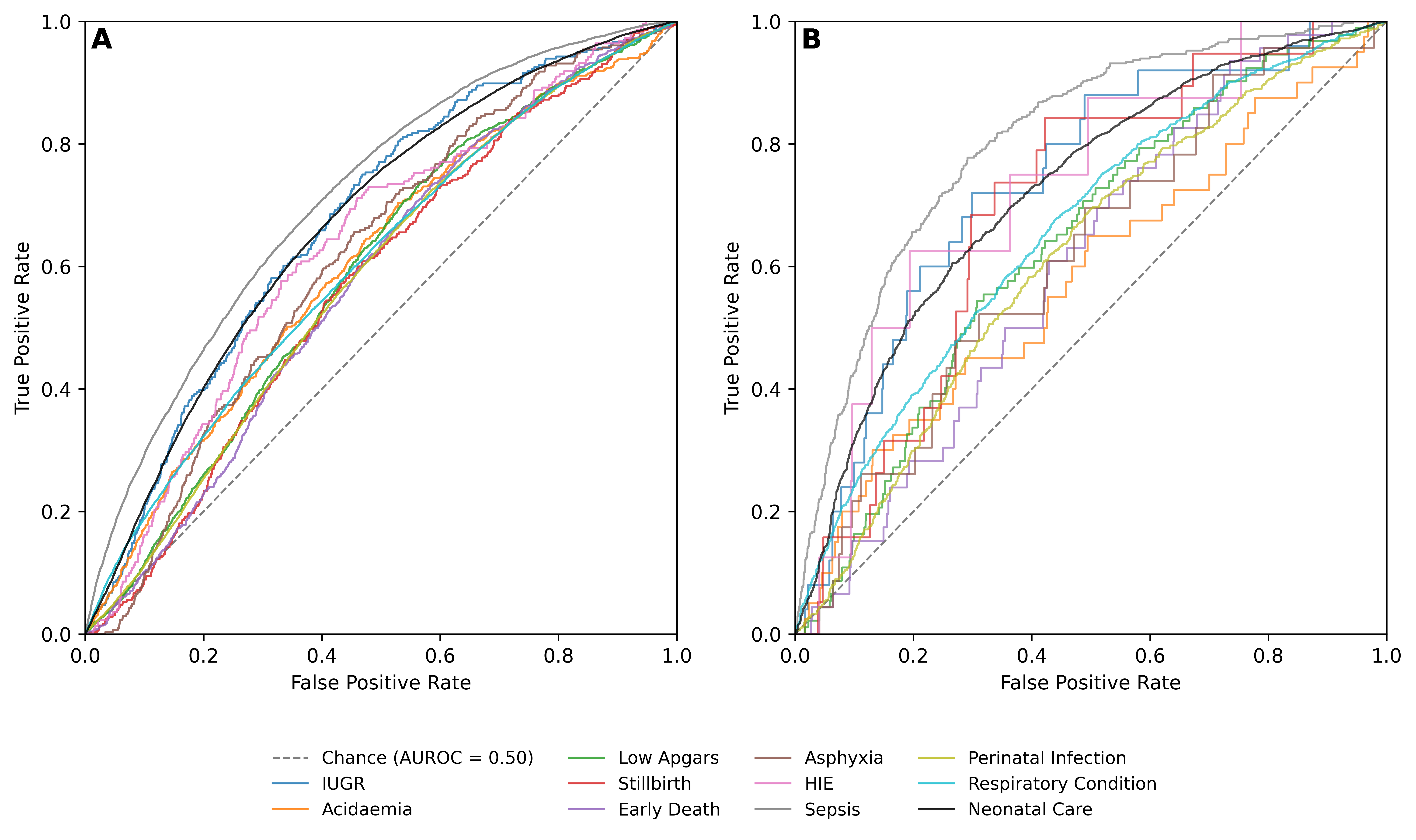}
    \caption{Receiver operating characteristic (ROC) curves for classifying specific conditions as adverse pregnancy outcomes (APOs) in the test dataset. Each line corresponds to a category of clinical conditions from the selection criteria for APOs as detailed in Figure \ref{fig:data-split-conditions}. \textbf{A} shows curves based on labels and scores for each individual fetal heart rate segment, while \textbf{B} compares median scores for all overlapping segments of complete cardiotocography signals with their labels. The dashed diagonal line represents the ROC curve for random selection of labels. The area under these curves (AUROCs) are presented in Table \ref{tab:condition_aucs}. IUGR = Intrauterine growth restriction. HIE = Hypoxic-ischaemic encephalopathy.}
    \label{fig:condition_predictions}
\end{figure}

\begin{table}[htbp]
    \centering
    \begin{tabular}{lcc}
        \toprule
        Condition & Segment AUROC & Case AUROC \\
        \midrule
        IUGR & 0.674 (0.646--0.702) & 0.735 (0.637--0.819) \\
        Acidaemia & 0.606 (0.583--0.629) & 0.580 (0.481--0.678) \\
        Low Apgars & 0.591 (0.576--0.604) & 0.636 (0.585--0.685) \\
        Stillbirth & 0.573 (0.548--0.599) & 0.694 (0.596--0.793) \\
        Early Death & 0.578 (0.559--0.596) & 0.602 (0.529--0.673) \\
        Asphyxia & 0.615 (0.587--0.642) & 0.616 (0.501--0.728) \\
        HIE & 0.634 (0.601--0.671) & 0.729 (0.553--0.889) \\
        Neonatal Sepsis & 0.713 (0.707--0.718) & 0.804 (0.782--0.824) \\
        Perinatal Infection & 0.582 (0.575--0.588) & 0.613 (0.588--0.637) \\
        Respiratory Condition & 0.606 (0.600--0.610) & 0.659 (0.639--0.678) \\
        Neonatal Care & 0.672 (0.667--0.676) & 0.721 (0.705--0.737) \\
        \bottomrule
    \end{tabular}
    \caption{Area under the receiver operating characteristic curves (AUROCs) for each group of conditions within the adverse pregnancy outcome group of the test dataset, as detailed in Figure \ref{fig:data-split-conditions}. Results are shown at the individual segment level and at the case level, median score across all overlapping segments for complete cardiotocographs. 95\% confidence intervals for each AUROC are included in brackets. IUGR = Intrauterine growth restriction. HIE = Hypoxic-ischaemic encephalopathy.}
    \label{tab:condition_aucs}
\end{table}

Figure \ref{fig:condition_predictions} presents ROC curves for each condition separately, comparing performance at both the segment level and the CTG level (by taking median predictions for all segments of each recording). The AUROC scores associated with each condition are presented with 95\% confidence intervals in Table \ref{tab:condition_aucs}. Segment-level predictions yielded variable AUROC values across conditions, ranging from 0.573 (stillbirth, 95\% CI: 0.548--0.599) to 0.713 (neonatal sepsis, 95\% CI: 0.707--0.718), with most conditions achieving AUROCs above 0.6. Aggregating scores at the CTG level generally improved discrimination, with the highest AUROC again observed for neonatal sepsis (0.804, 95\% CI: 0.782--0.824), and a strong improvement for classifying stillbirths (0.694, 95\% CI: 0.596--0.793).


\subsection{Latent Space Interpretation}

\begin{figure}[htbp]
    \centering
    \includegraphics[width=\linewidth]{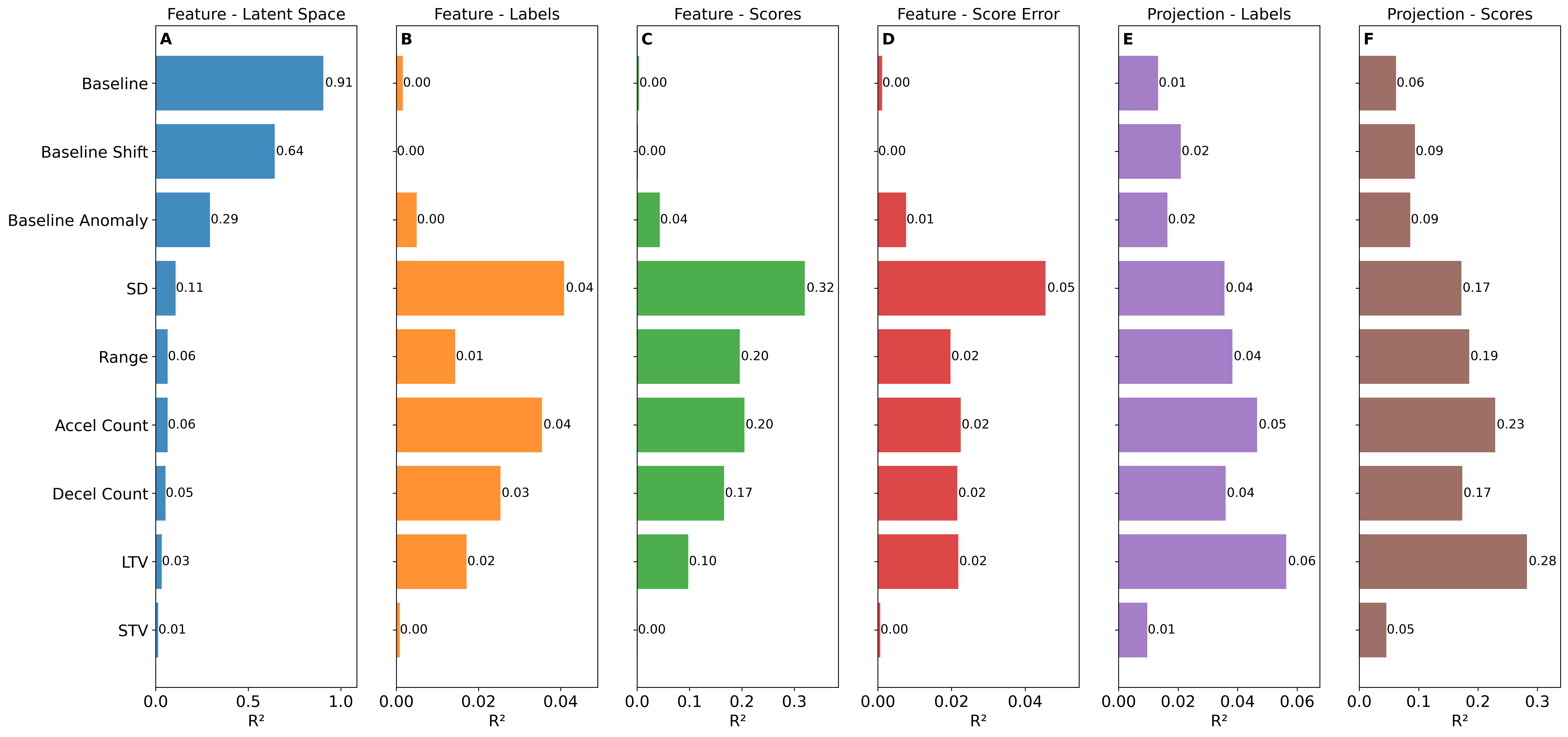}
    \caption{Coefficients of determination ($R^2$) between extracted features of five-minute fetal heart rate segments from the test dataset and various aspects of the model. Panel \textbf{A} shows $R^2$ values between features and their corresponding latent space representations. Panels \textbf{B}, \textbf{C}, and \textbf{D} show $R^2$ values between features and labels, scores and prediction error, respectively. Panels \textbf{E} and \textbf{F} present $R^2$ values between labels and scores, respectively, and the projections of latent variables along the direction in latent space most correlated with each feature. SD = standard deviation. LTV = long-term variability. STV = short-term variability.}
    \label{fig:feature_r2_scores}
\end{figure}

Figure~\ref{fig:feature_r2_scores} presents the coefficient of determination ($R^2$) between extracted input features and various components of the model and dataset. Panel~\textbf{A} shows that features such as the baseline heart rate ($R^2 = 0.91$) and the baseline shift over time ($R^2 = 0.64$) are strongly represented in the latent space, suggesting these features are primary drivers of the learned representation. The deviation from the overall mean baseline (baseline anomaly) also shows a moderate correspondence ($R^2 = 0.29$). Other features, such as variability metrics and deceleration counts, exhibit limited alignment with the latent space ($R^2 < 0.1$).

Panel~\textbf{B} shows that the correlation between features and labels is weak overall ($R^2 < 0.05$), whereas panel~\textbf{C} indicates stronger associations with scores. Notably, the segment standard deviation, which is weakly encoded in the latent space, shows the highest $R^2$ with scores ($R^2 = 0.32$). Panel~\textbf{D} displays the $R^2$ between features and error between scores and labels, revealing no substantial association across features, implying that individual features are not directly responsible for model inaccuracy. Panels~\textbf{E} and~\textbf{F} assess the extent to which projections of latent variables, along directions most correlated with each feature, align with labels and scores. These show modest correlations (up to $R^2 = 0.28$), indicating that while some label-relevant information is embedded in the latent space, it is dispersed and entangled across multiple dimensions.

\begin{figure}[htbp]
    \centering
    \includegraphics[width=\linewidth]{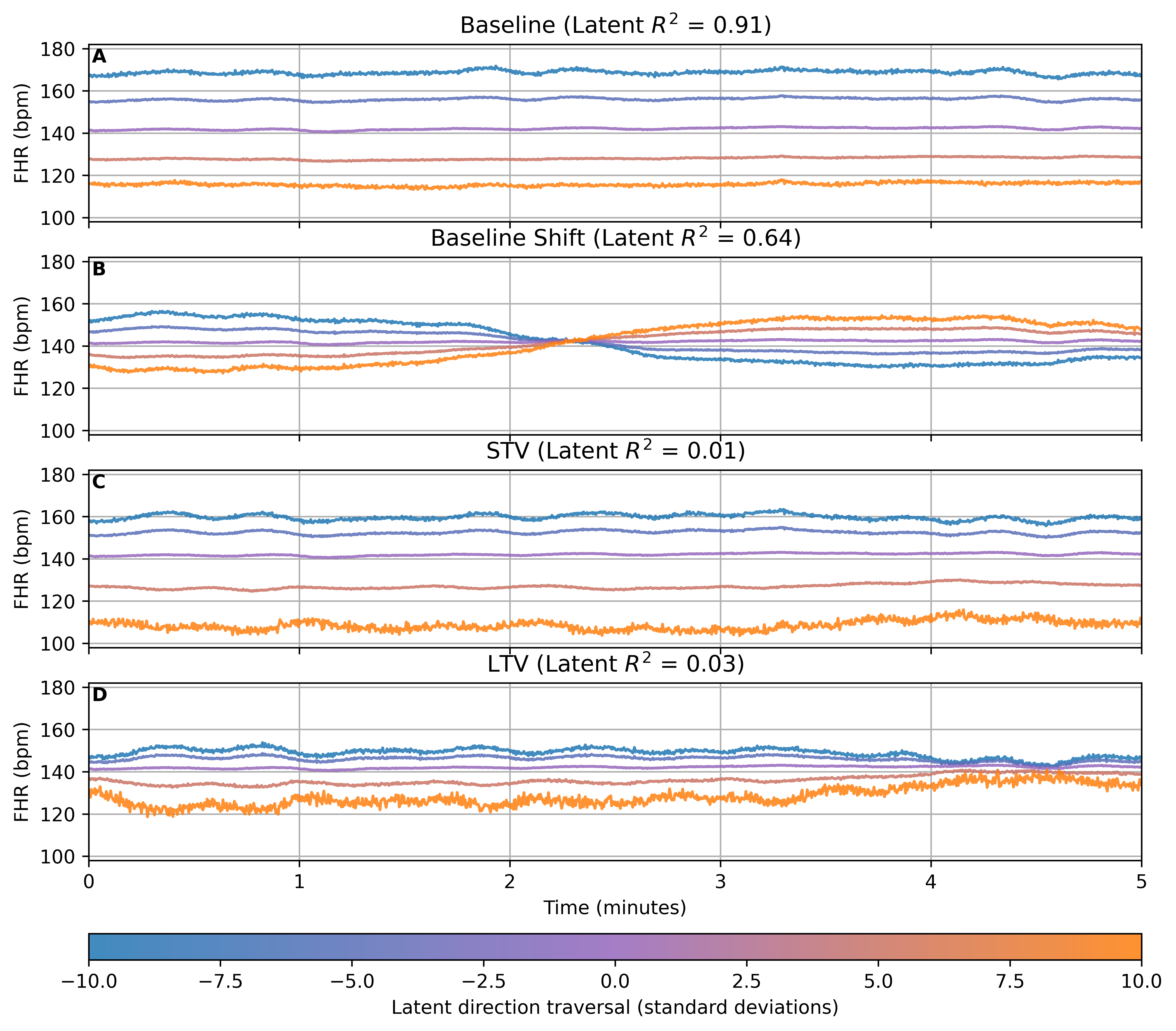}
    \caption{Latent traversals along directions most correlated with selected fetal heart rate features. Each panel shows decoded signals as the mean latent representation is perturbed in the direction associated with one feature (from $-10$ to $+10$ standard deviations). Panel~\textbf{A} (baseline fetal heart rate) and \textbf{B} (baseline shift over time) show clean changes aligned with the semantic meaning of the features, while Panel~\textbf{C} (short-term variability, STV) and Panel~\textbf{D} (long-term variability, LTV) show minimal association, reflecting their weak latent encoding and entanglement with dominant latent directions.}
    \label{fig:feature_traversals}
\end{figure}

Figure~\ref{fig:feature_traversals} illustrates latent traversals for four selected features: baseline, baseline shift, short-term variability (STV), and long-term variability (LTV). Each panel shows how decoded FHR signals change as the latent variables are perturbed along directions most correlated with a given feature. Baseline (Panel~\textbf{A}) and baseline shift (Panel~\textbf{B}) exhibit clean, interpretable changes: vertical translation and diagonal drift, respectively. STV (Panel~\textbf{C}) produces increased variability in the positive direction but also modulates the baseline, suggesting entanglement with dominant latent directions. LTV (Panel~\textbf{D}) induces similar but minimal signal changes, consistent with its low latent $R^2$.

\begin{figure}[htbp]
    \centering
    \includegraphics[width=\linewidth]{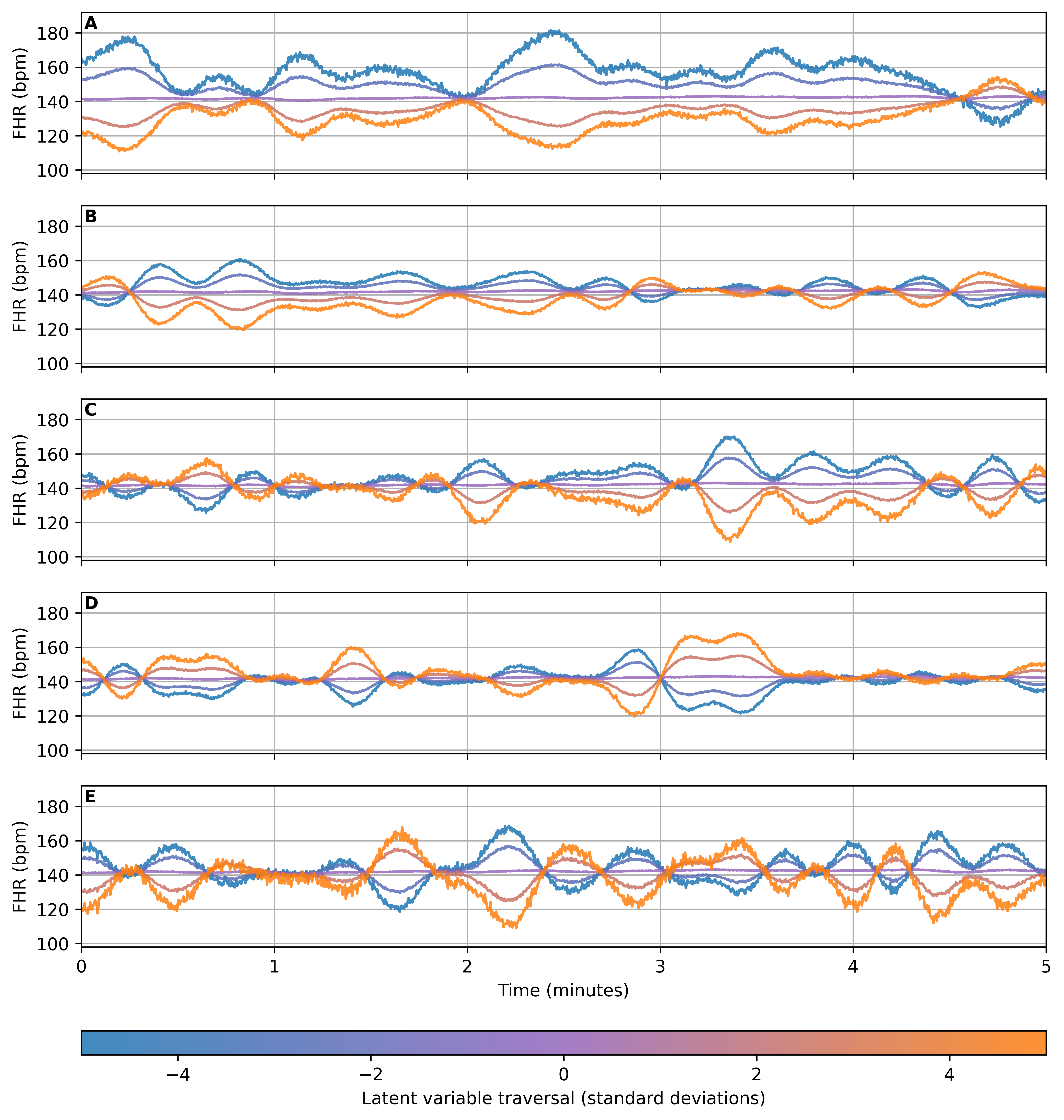}
    \caption{Traversals along five individual latent dimensions, showing decoded fetal heart rate signals as the latent mean is perturbed along each (from $-5$ to $+5$ standard deviations). Panel~\textbf{A} shows traversals for the dimension most strongly associated with baseline, and Panel~\textbf{B} for the dimension most strongly associated with baseline shift. Panel~\textbf{C} shows traversals for another dimension linked to baseline features. Panels~\textbf{D} and~\textbf{E} illustrate typical latent dimension traversals that add pseudo-sinusoidal signals with little interpretable meaning, but which in combination can generate the diverse shapes and features observed in the input signals.}
    \label{fig:individual_traversals}
\end{figure}

To further understand the structure of the learned latent space, decoded signals resulting from traversals along individual latent dimensions were examined (see examples in Figure~\ref{fig:individual_traversals}). The latent variable shifted in Panel~\textbf{A} is the dominant contributor to the direction most strongly associated with baseline and produces clear vertical shifts in the decoded signal. Similarly, the variable in Panel~\textbf{B}, which contributes most to the direction aligned with baseline shift, induces a subtle effect of either increasing or decreasing over time, depending on the direction of traversal. A few other dimensions showed weaker but still discernible effects. For example, Panel~\textbf{C} exhibits a mild relationship with baseline, with signals tending to rise in one direction and fall in the other. However, most latent variables exhibited weakly structured or chaotic signal changes that lacked clear alignment with clinically meaningful features. In many cases, pseudo-sinusoidal patterns appeared without clear localisation or interpretability, such as the dimensions traversed in Panels \textbf{D} and \textbf{E}. An unsupervised analysis of the latent space revealed similar patterns. The dominant principal and independent components were primarily aligned with baseline levels and baseline shifts, while other components captured minimal, more irregular or chaotic variations in the mean latent signal.


\section{Discussion}

\subsection{Model Assessment and Utility}

This study demonstrates the use of a supervised variational autoencoder (VAE) to classify cardiotocography (CTG) signals according to fetal outcomes. The model discriminated between segments from normal pregnancy outcome (NPO) and adverse pregnancy outcome (APO) cases with moderate accuracy (AUROC = 0.75, sensitivity = 83.7\%, specificity = 71.9\%), which improved when predictions were aggregated across entire CTG recordings (AUROC = 0.78, sensitivity = 82.6\%, specificity = 82.6\%). This improvement reflects the benefit of utilising longer durations of CTG data to capture a broader range of features relevant to outcome classification. Aggregating predictions across the full CTG improved overall discrimination, particularly specificity, supporting more reliable outcome assessment in a clinical context. This benefit comes with a trade-off, as aggregation would require longer CTG recordings before predictions could be made.

Our model performed with comparable accuracy to other antepartum outcome prediction models. PatchCTG, a transformer-based model from our group has previously achieved AUROC of 0.77, sensitivity of 57\% and specificity of 88\% when predicting adverse outcomes, while a model predicting early-onset neonatal sepsis after preterm, prelabor rupture of membranes achieved AUROC of 0.734 \cite{khan2024patchctg, birgisdottir2025riskmodel}. The Dawes–Redman system is widely used in clinical practice to identify fetal well-being antepartum and demonstrates high specificity (90.7\%; 95\% CI, 89.2–92.0\%) but low sensitivity (18.2\%; 95\% CI, 16.3–20.0\%) for ruling out adverse pregnancy outcomes \cite{jones2025drperformance}. In contrast, the model developed in this study achieves substantially higher sensitivity while maintaining reasonable specificity, highlighting its potential to better identify at-risk pregnancies. Importantly, the VAE provides more interpretable representations of CTG signals than other models, allowing visualisation of features driving predictions and offering a potential basis for future clinical use.

Performance varied across outcome categories, with the greatest gains from aggregation observed for stillbirth, neonatal sepsis and hypoxic-ischaemic encephalopathy, suggesting that some conditions are more readily captured through broader temporal trends rather than isolated features. Conversely, the model was less effective at detecting acidemia when scores were aggregated, likely because acidemia often develops acutely during labour and is therefore less likely to be apparent in antepartum CTG patterns. These findings underscore the importance of considering temporal context in automated CTG classification. Rather than relying on short-duration signal segments, models that incorporate or summarise information over longer windows may better capture subtle or evolving physiological patterns associated with some APOs. This approach is especially relevant for retrospective or diagnostic classification, and may also inform decision support tools in obstetric care.

Although the model was trained using outcome-level labels, it appears capable of identifying segments within otherwise normal CTG traces that exhibit clinically non-reassuring patterns. This suggests that the VAE’s reconstruction and scoring may capture localised deviations in FHR variability or accelerations that are not evident from aggregate outcome labels alone. Such sensitivity can enhance interpretability and could potentially support clinical decision-making by highlighting periods warranting closer attention, even in generally reassuring recordings. However, these segment-level predictions are exploratory and should be interpreted with caution, since the model was not explicitly trained to classify individual segments.

Our investigation of latent structure using total correlation (TC) constraints suggests that models with higher TC values (i.e., less restricted disentanglement) perform better in terms of both reconstruction error and classification accuracy. Setting a high TC target led to an effective $\lambda = 0$, allowing the model to use all latent dimensions freely. This observation supports the hypothesis that strict enforcement of disentanglement may impair task-relevant feature encoding in this setting, especially when the clinical signal is subtle or complex \cite{harvey2024comparison}.

\subsection{Model Interpretability}

The interpretability analyses reveal that baseline-related features (e.g., baseline FHR and baseline shift) are strongly encoded in the latent space and align well with model predictions. These features are known to have clinical relevance and are prominent in human CTG assessment guidelines. In contrast, features such as short- and long-term variability (STV and LTV, respectively) showed weaker representation in the latent space and limited correlation with model predictions. While these features are clinically informative they describe chaotic fluctuations over time, making it less likely for the model to associate with specific patterns. However, latent traversals along the directions most associated with STV and LTV suggest that the model has captured the known inverse relationship between baseline and variability \cite{bhide18baselinestvrel}. Only a subset of features produce clean, directional changes in decoded signals, while many dimensions contribute non-linearly or in entangled ways.

Deep learning models trained on more periodic physiological signals often uncover cleaner and more interpretable latent features. For example, in studies using VAEs on electrocardiogram (ECG) data, latent dimensions have been shown to correspond to distinct waveform components such as heart rate, QRS axis, or modulation of P, R, and T wave amplitudes and durations \cite{jang21unsupervised, beetz22multidomain, patika24artificial}. This level of interpretability is supported by the fact that ECG waveforms arise from a relatively simple physiological process. In contrast, fetal heart rate signals are irregular and shaped by a complex, often chaotic interplay of fetal, placental and maternal processes. They lack a consistent or cyclical form and vary over both short and long timescales, which makes it difficult for the model to disentangle distinct physiological factors. This likely explains why only a subset of latent dimensions in our model aligned with interpretable features, while many others contributed in more entangled or non-linear ways.

Interestingly, some features with weak latent alignment (such as segment standard deviation) still correlated strongly with the predicted scores. This suggests that the model may learn to use complex feature combinations that do not map cleanly to human-interpretable metrics but are still predictive of outcomes. Similarly, although no single latent dimension was strongly associated with accelerations or decelerations, their reconstruction required coordinated changes across multiple latent dimensions, implying a distributed and entangled representation. The stronger correlation between predicted scores and standard deviation (compared to STV or LTV) may reflect the fact that it captures variability across the entire segment, including regions containing accelerations and decelerations, which are intentionally excluded in the calculation of STV and LTV. These findings underscore the challenge of aligning learned representations with clinically defined features and the potential for discovering new latent biomarkers.

\subsection{Limitations and Future Work}

There are several limitations to this work. Since labels were based on postnatal outcomes rather than expert CTG assessments, some parts of the signal that experts might consider suspicious or pathological could be labeled as normal if the fetus compensated well or if they eventually led to interventions preventing adverse pregnancy outcomes.

It should be acknowledged that certain conditions such as acidemia often arise as a consequence of intrapartum events. While analysis of antenatal FHR data may reveal patterns indicative of underlying fetal vulnerability and a higher likelihood of compromised tolerance to labour, some pathologies develop due to acute events during labour itself, for which antepartum assessments may show no detectable association.

The selected APO cases were all preterm, due to the restriction that CTGs be recorded before 37 weeks. While this restriction was intended to avoid late-pregnancy effects, it introduces prematurity as a potential confounding factor in the association between CTG signals and APOs.

Although some latent dimensions align well with clinically meaningful features, many do not. This limits transparency and interpretability, which may reduce the suitability of such a model in settings where understanding the basis of predictions is essential, such as real-time clinical decision-making or regulatory contexts. Nonetheless, the model still supports clinically useful prediction of APOs and offers a foundation for future work on interpretable deep learning models for CTG analysis.

Future work might aim to validate this approach on external datasets, incorporate richer modalities (e.g., uterine contraction signals, maternal heart rate), test alignment with expert human annotations or train similar models on more specific APOs. Additionally, the latent structure uncovered here may offer a foundation for semi-supervised or unsupervised discovery of fetal distress subtypes, potentially informing future clinical decision-support tools.


\section{Conclusions}

A supervised variational autoencoder was developed for the analysis of fetal heart rate signals, trained on outcome-labeled cardiotocography (CTG) segments from the OxMat dataset. The model achieved moderate discriminative ability while learning a structured latent space that encodes clinically relevant features such as baseline heart rate and its change over time. Although disentanglement was limited, interpretability analyses revealed meaningful directions in the latent space aligned with known physiological patterns. These findings highlight the importance of incorporating temporal context for improved prediction and support the potential of probabilistic deep generative models for interpretable fetal outcome prediction. This approach may serve as a foundation for discovering novel physiological markers associated with adverse pregnancy outcomes and inform future clinical decision-support tools focused on outcome assessment, despite current challenges in full latent feature disentanglement.

\section{Author Contributions}

John Tolladay: Conceptualisation, data curation, formal analysis, investigation, methodology, software, resources, validation, visualisation, writing - original draft, writing - review editing. Beth Albert: Project administration, writing - review \& editing. Gabriel Davis Jones: Conceptualization, funding acquisition, project administration, resources,
supervision, validation, writing - review \& editing.

\section{Declarations}

\noindent\textbf{Ethics:} This study was approved by the Ethics Committee in the Joint Research Office, Research and Development Department, Oxford University Hospitals NHS Trust (approval number: 25/HRA/1966, granted on 13th May 2025)

\noindent\textbf{Funding:} This study was supported by the Medical Research Council (UKRI grant MR/X029689/1).

\noindent\textbf{Data availability:} The code and anonymised data used in this study may be made available upon reasonable request from the corresponding author, subject to ethics approval.

\noindent\textbf{Conflicts of interest:} The authors declare no conflicts of interest.

\bibliography{references}

\end{document}